\documentclass{article}
\usepackage{spconf,amsmath,epsfig}
\usepackage{cite}
\usepackage{graphicx}
\graphicspath{{Figures/}}
\DeclareGraphicsExtensions{.pdf,.jpeg,.png}
\usepackage{algorithm}
\usepackage{amsmath}
\usepackage{algorithmic}
\usepackage{gensymb}
\usepackage{subfloat}
\usepackage{array}
\usepackage{multirow}
\usepackage{amsfonts}
\usepackage{stfloats}
\usepackage{url}
\usepackage[export]{adjustbox}
\usepackage{textcomp}
\usepackage{cite}

\usepackage{caption}
\usepackage[font=footnotesize]{subcaption}
\usepackage{breqn}
\usepackage{tikz}
\usepackage{pgfplots}
\usetikzlibrary{shapes, positioning, matrix, calc, pgfplots.statistics,
pgfplots.groupplots, pgfplots.colorbrewer, 3d, arrows, positioning, quotes,
backgrounds, arrows.meta, spy}
\input{tikz_nn.tex}

\begin{document}

\title{Weakly supervised marine animal detection from remote sensing images using vector-quantized variational autoencoder}
%
\name{Minh-Tan Pham $^*$, Hugo Gangloff $^{*,\dagger}$, Sébastien Lefèvre $^*$  \thanks{This work was done in the context of the SEMMACAPE project, which benefits from an ADEME ({\em Agence de la transition écologique}) grant under the “Sustainable Energies” call for research projects (2018--2019).}}

\address{ $^*$ Université Bretagne Sud, UMR IRISA 6074, 56000 Vannes, France \\
		$^{\dagger}$ INRAE, AgroParisTech, UMR MIA Paris-Saclay, 91120 Palaiseau, France\\
		\texttt{\ninept minh-tan.pham@irisa.fr, hugo.gangloff@inrae.fr, sebastien.lefevre@irisa.fr}}

	\maketitle
	
\begin{abstract}
This paper studies a reconstruction-based approach for weakly-supervised animal detection from aerial images in marine environments. Such an approach leverages an anomaly detection framework that computes metrics directly on the input space, enhancing interpretability and anomaly localization compared to feature embedding methods. Building upon the success of Vector-Quantized Variational Autoencoders in anomaly detection on computer vision datasets, we adapt them to the marine animal detection domain and address the challenge of handling noisy data.
To evaluate our approach, we compare it with existing methods in the context of marine animal detection from aerial image data. Experiments conducted on two dedicated datasets demonstrate the superior performance of the proposed method over recent studies in the literature. Our framework offers improved interpretability and localization of anomalies, providing valuable insights for monitoring marine ecosystems and mitigating the impact of human activities on marine animals.
\end{abstract}
	
	\begin{keywords}
	Anomaly detection, weakly-supervised learning, aerial remote sensing, marine animal monitoring
	\end{keywords}

\section{Introduction}
With the increasing exploitation of marine resources nowadays, monitoring human activities in the sea environments becomes crucial since they could have a significant impact on marine animals \cite{bergstrom2014effects}. To prevent negative influences from these activities to the ecosystem, aerial surveys are exploited to capture and detect the presence of marine animals for monitoring their population and behavior. However, the visual analysis of such aerial images and videos is very time-consuming and often requires expert knowledge. The emerging development of deep learning techniques such as Convolutional Neural Networks (CNNs) enables us to automatically perform this task. Nevertheless, most of the effective methods using CNNs for animal detection from aerial images are conducted using supervised learning \cite{maire2015automating, boudaoud2019marine}. To reduce the high cost of data annotation, recent studies have investigated weakly-supervised approaches such as \cite{fasana2022weakly, berg2022weakly}. In these studies, only image-level annotations (which are faster and cheaper to obtain) are used for training instead of object-level labels (bounding boxes) required by fully supervised frameworks.
In \cite{berg2022weakly}, the authors formulate the weakly-supervised animal detection as an anomaly detection (AD) task where deep networks are first trained only on empty images (i.e., normal data), while the prediction detect animals as anomalies. To do so, they exploit the CNN feature embedding to model the statistical distribution of normal data and compute the anomaly score using distribution-based distance metrics. Compared to previous studies of AD on natural images \cite{defard2021padim, kim2021semi}, the adaptations and improvements proposed in \cite{berg2022weakly} make this approach more suitable for remote sensing data, as proved by its higher detection performance.

In this work, we adopt another anomaly detection framework using a reconstruction-based approach to perform weakly-supervised animal detection from aerial images. This framework is appealing given it relies on metrics computed on the input space of images. As such, it could provide easier interpretability and localization of the anomalies than feature embedding-based approach \cite{ruff2019deep}. Indeed, we build upon our recent work on VQ-VAEs (Vector-Quantized Variational Autoencoders) \cite{gangloff2022leveraging} that successfully performed AD on standard computer vision datasets. Here, we investigate its behavior and perform a comparison with respect to literature works in the context of marine animal detection from remote sensing data. We also propose some adaptations of the VQ-VAE framework to remain efficient on noisy data. Experimental results on two dedicated datasets show better performance than feature embedding methods proposed recently in \cite{berg2022weakly, defard2021padim, kim2021semi}.

In the rest of this paper, Sec. \ref{sec:2-Methodology} overviews the the VQ-VAE framework proposed in \cite{gangloff2022leveraging} for AD and provides our improvements to deal with noise in images using the hysteresis thresholding technique \cite{angulo2003morphological}. Sec. \ref{sec:3-ExpResults} describes our experiments conducted on two aerial image datasets used in \cite{berg2022weakly} in the context of weakly supervised marine animal detection. We draw some conclusions and potential future works in Sec. \ref{sec:4-Conclusions}.
\section{Methodology}
\label{sec:2-Methodology}
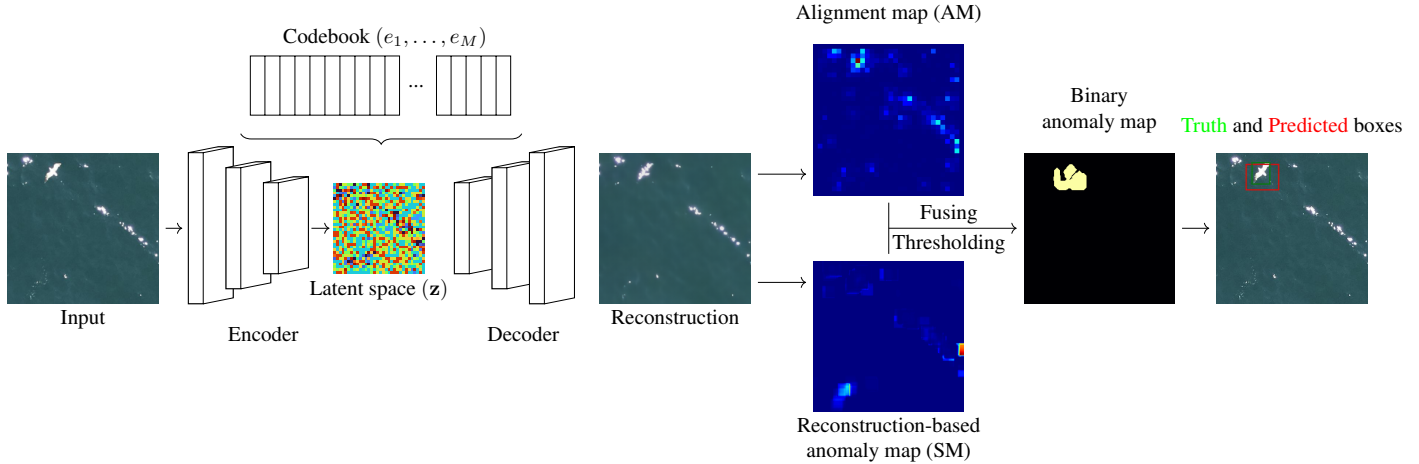
\begin{figure*}[htb]
    \centering
\scalebox{0.8}{
\begin{tikzpicture}
    \node[label={[below,xshift=0em, yshift=-2.6cm] Input}] (input) at (0,0) {\includegraphics[width=2.5cm]{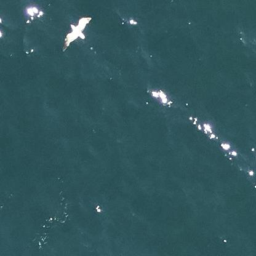}};
    \node[plate={0.1cm}{2.5cm}{0.5cm},
    right = 0.5cm of input,
    anchor=center,
    rotate=0] (enc1) {};
    \node[plate={0.1cm}{2cm}{0.5cm},
    right = 0.5cm of enc1,
    anchor=center,
    rotate=0] (enc2) {};
    \node[plate={0.1cm}{1.5cm}{0.5cm},
    right = 0.5cm of enc2,
    anchor=center,
    rotate=0] (enc3) {};
    \node[below = 0.5cm of enc2, xshift=0.5cm] (encoder) {Encoder};
    \node[right = 0.8cm of enc3, label={[below,xshift=0em, yshift=-1.6cm] Latent space $(\mathbf{z})$}] (ls) {\includegraphics[width=1.5cm]{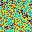}};
    
    \draw[->] (input.east) -- ($(enc1.west)+(-1mm, 0mm)$);
    \draw[->] ($(enc3.east)+(5.5mm,0mm)$) -- ($(ls.west)+(0.5mm, 0mm)$);
    
    \node[label={[below,xshift=1cm, yshift=0.6cm] Codebook $(e_1,\dots,e_M)$},
    above = 1cm of ls, xshift=-0.9cm,
    rectangle split,
    rectangle split horizontal,
    rectangle split every empty part={},
    rectangle split empty part height=0.75cm,
    rectangle split empty part width=0.001cm,
    rectangle split parts=10,
    draw, node distance=1em] (cb) {};
    \node[label={[left,xshift=-0.7cm, yshift=-0.5cm]...},
    right = 0.6cm of cb,
    rectangle split,
    rectangle split horizontal,
    rectangle split every empty part={},
    rectangle split empty part height=0.75cm,
    rectangle split empty part width=0.001cm,
    rectangle split parts=5,
    draw, node distance=1em] (cb2) {};
    \draw [decorate, decoration={brace, mirror, amplitude=5pt}] ($(cb)+(-14mm,-8mm)$) -- ($(cb2)+(8mm,-8mm)$) node [below, black, midway, yshift=-0.1cm, text width=2cm, align=center] {};
    
    \node[plate={0.1cm}{1.5cm}{0.5cm},
    right = 0.5cm of ls,
    anchor=center,
    rotate=0] (dec1) {};
    \node[plate={0.1cm}{2cm}{0.5cm},
    right = 0.5cm of dec1,
    anchor=center,
    rotate=0] (dec2) {};
    \node[plate={0.1cm}{2.5cm}{0.5cm},
    right = 0.5cm of dec2,
    anchor=center,
    rotate=0] (dec3) {};
    \node[below = 0.5cm of dec2, xshift=0.4cm] (decoder) {Decoder};
    \node[right = 0.8cm of dec3, label={[below,xshift=0em, yshift=-2.6cm] Reconstruction}] (rec) {\includegraphics[width=2.5cm]{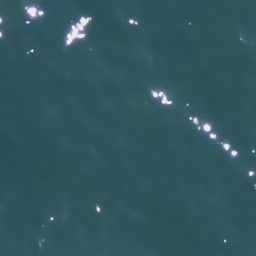}};

    \node[right = 0.8cm of rec, yshift=1.8cm, label={[above,xshift=0em, yshift=0.1cm] Alignment map (AM)}] (am) {\includegraphics[width=2.5cm]{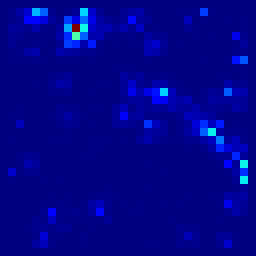}};

    \node[right = 0.8cm of rec, yshift=-1.8cm, label={[below,xshift=0em, yshift=-2.6cm, align=center] Reconstruction-based \\ anomaly map (SM)}] (sm) {\includegraphics[width=2.5cm]{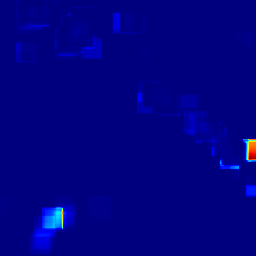}};

     \node[right = 4.3cm of rec, label={[above,xshift=0em, yshift=0.1cm, align=center] Binary \\ anomaly map}] (amap) {\includegraphics[width=2.5cm]{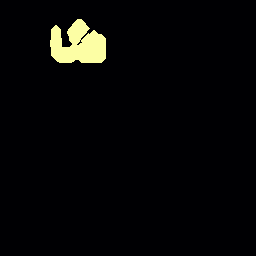}};

     \node[right = 7.5cm of rec, label={[above,xshift=0em, yshift=0.1cm, align=center] \textcolor{green}{Truth} and \textcolor{red}{Predicted} boxes}] (gt) {\includegraphics[width=2.5cm]{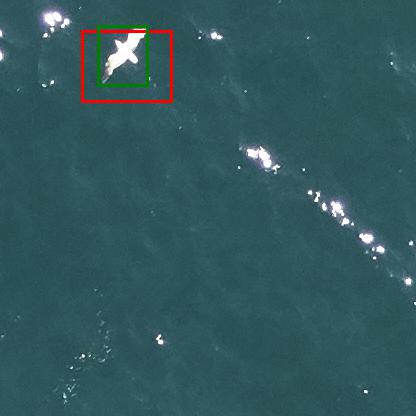}};

    \draw[->] ($(rec.east)+(0mm,9mm)$) -- ($(am.west)+(0mm, -9mm)$);

    \draw[->] ($(rec.east)+(0mm,-9mm)$) -- ($(sm.west)+(0mm, 9mm)$);
    \draw ($(am.south)+(0mm,0mm)$) -- ($(sm.north)+(0mm, 0mm)$);
    \draw[->] ($(am.south)+(0mm,-4.25mm)$) -- ($(amap.west)+(0mm, 0mm)$);

    \node[below = -1mm of am,xshift=10mm] (fusing) {Fusing};
    \node[below = 4mm of am,xshift=10mm] (th) {Thresholding};
    \draw[->] ($(amap.east)+(0mm,0mm)$) -- ($(gt.west)+(0mm, 0mm)$);
    
\end{tikzpicture}
}
\caption{Overview of the proposed AD framework using VQ-VAE. To deal with noisy data in the context of marine animal detection, both AM and SM are thresholded before being fused to produce the final anomaly map (see Algorithm~\ref{algo:vqvae_noisy}). }
    \label{fig:workflow}
\end{figure*}

\subsection{VQ-VAEs for anomaly detection}
Reconstruction-based AD methods compute anomaly score based on the distance between the input image and its reconstructed version given by a deep network. VAEs are the most popular models that have been adopted within this context. First introduced in \cite{van2017neural}, VQ-VAEs with discrete latent variables have been recently shown to provide more accurate reconstructions than standard VAEs. This makes VQ-VAEs promising models for reconstruction-based AD, since they could result in less noisy residual images. This higher performance has been proved in \cite{gangloff2022leveraging} on different vision image benchmarks. Fig. \ref{fig:workflow} illustrates the workflow of VQ-VAEs which is quite similar to classical VAEs, apart from their discrete latent space linked to a particular codebook vector. To compute the final anomaly map, \cite{gangloff2022leveraging} proposed to fuse two components including the reconstruction-based anomaly map (denoted by $SM$) computed using the Struture Similarity Index Measure (SSIM), and the alignment map ($AM$) computed from the latent space of the VQ-VAE. $AM$ is then upsampled to the image scale based on morphological dilation operation and fused with the $SM$ to produce the final anomaly map. For more details about the computations and fusion of these two metrics, we refer readers to \cite{gangloff2022leveraging}. To this end, the principle of measuring anomaly score based on both reconstruction and discrete latent space of the model makes the VQ-VAE different and more effective than standard VAE-based AD models.

\subsection{Adaptations to deal with noisy data}
\label{subsec:adapt}
\begin{algorithm}[h]
{\bf Input:} 
\begin{itemize}
\itemsep-0.2em
    \item $SM$: the SSIM anomaly map
    \item $AM$: the alignment anomaly map
    \item $\lambda_{SM}$, $\lambda_{AM}$: thresholds for $SM$ and $AM$ segmentation
\end{itemize}

{\bf Output}: $Amap$: the final binary  anomaly map
\begin{enumerate}
\itemsep-0.2em
    \item Segment $SM$ at threshold  $\lambda_{SM}$ $\rightarrow$ $\overline{\rm SM}$
    \item Segment $AM$ at threshold  $\lambda_{AM}$  $\rightarrow$ $\overline{\rm AM}$
    \item Get the connected components of $\overline{\rm SM}$ $\rightarrow$ $\widetilde{\rm SM}$
    \item Keep the connected components of $\widetilde{\rm SM}$ which have non null intersection with $\overline{\rm AM}$ $\rightarrow$ ${\rm Amap}$
\end{enumerate}
\caption{Using the VQ-VAE for AD in noisy data.}
\label{algo:vqvae_noisy}
\end{algorithm}
Similar to most applications of deep frameworks developed on natural images into the remote sensing field, the use of VQ-VAEs to detect marine animals from aerial images is not direct and requires some adaptations to consider the data particularities. Indeed, aerial images acquired over the sea surface for marine monitoring often face with some issues:
\begin{itemize}
    \item The sun glare which saturates many images makes it difficult to discriminate from the birds which we want to detect, thus tends to induce many false positives.
    \item The appearance of underwater animals (dolphin, porpoise, etc.) is very varied, due to their positions from the sea surface. Moreover, their movements also induce foam that makes the image very noisy.
\end{itemize}
To address these complex scenarios, we propose a dedicated approach using thresholding by hysteresis technique~\cite{angulo2003morphological} (i.e. double thresholding) that uses the segmented $AM$ as markers to keep connected components of interest from the segmented $SM$. Such an approach exploits the $AM$ as a way to discard false positive results from the $SM$. The proposed method is summarized in Algorithm~\ref{algo:vqvae_noisy}. 

\section{Experiments}
\label{sec:3-ExpResults}
\subsection{Datasets and setup}
Our experiments are conducted on two dedicated marine animal image datasets which are called \emph{Semmacape} and  \emph{Kelonia}, recently introduced and studied in \cite{berg2022weakly}. We adopt the similar data splits to perform AD for weakly-supervised animal detection: only normal data (images without animals) are used for training. The proposed approach is then compared against three recent methods including PaDIM \cite{defard2021padim}, OrthoAD \cite{kim2021semi} and PaDiM+NF \cite{berg2022weakly}. To conduct experiment on Semmacape data which are highly noisy, our proposed adaptation in Sec. \ref{subsec:adapt} is adopted. For model and training setup, we basically follow the setting of the standard VQ-VAE architecture in \cite{van2017neural,gangloff2022leveraging}. For the three reference methods, we also adopt the similar setting as done in \cite{berg2022weakly}.

\subsection{Results and analysis}
\begin{figure*}[h]
\centering
\makebox[0.9\textwidth][c]{
\begin{tikzpicture}
    \node[label={[above,xshift=0em, yshift=-0.08cm]Input}] (a){\includegraphics[width=2.5cm]{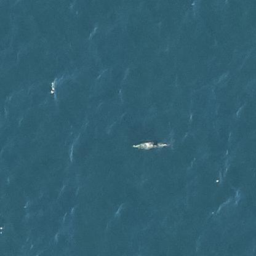}}; 
    \node[label={[above,xshift=0em, yshift=0cm]Reconstruction}](b) at (a.east) [xshift=11.2mm]{\includegraphics[width=2.5cm]{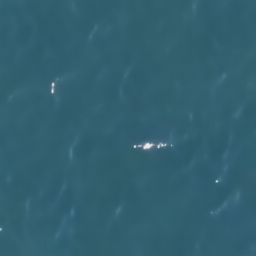}}; 
    \node[label={[above,xshift=0em, yshift=0cm]AM}](c) at (b.east) [xshift=11.2mm]{\includegraphics[width=2.5cm]{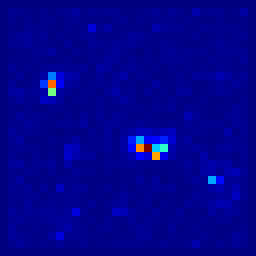}}; 
    \node[label={[above,xshift=0em, yshift=0cm]SM}](d) at (c.east) [xshift=11.2mm]{\includegraphics[width=2.5cm]{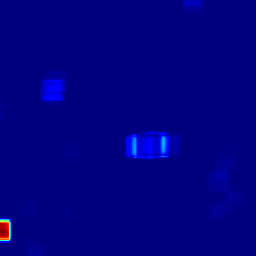}}; 
    \node[label={[above,xshift=0em, yshift=-0.08cm]Amap}](e) at (d.east) [xshift=11.2mm]{\includegraphics[width=2.5cm]{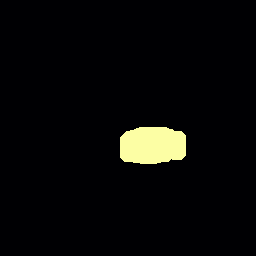}}; 
    \node[label={[above,xshift=0em, yshift=0cm]{\color{green}GT}  and {\color{red}Pred.}}](f) at (e.east) [xshift=11.2mm]{\includegraphics[width=2.5cm]{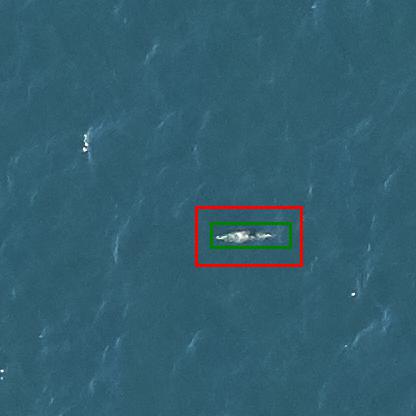}}; 
    \node at (a.south)[yshift=-11.5mm] {\includegraphics[width=2.5cm]{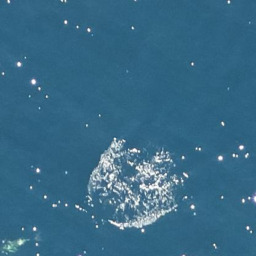}}; 
    \node at (b.south)[yshift=-11.5mm] {\includegraphics[width=2.5cm]{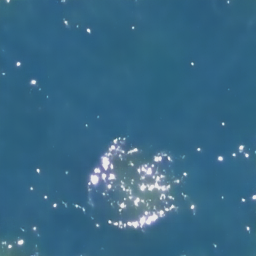}}; 
    \node at (c.south)[yshift=-11.5mm]{\includegraphics[width=2.5cm]{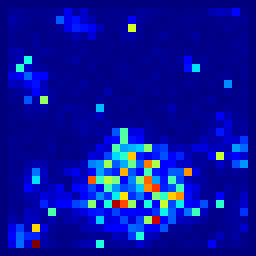}}; 
    \node at (d.south)[yshift=-11.5mm]{\includegraphics[width=2.5cm]{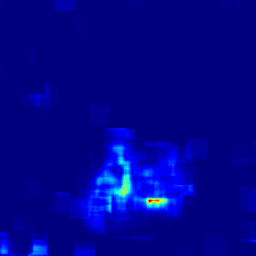}}; 
    \node at (e.south)[yshift=-11.5mm]{\includegraphics[width=2.5cm]{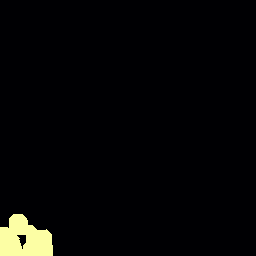}}; 
    \node at (f.south)[yshift=-11.5mm]{\includegraphics[width=2.5cm]{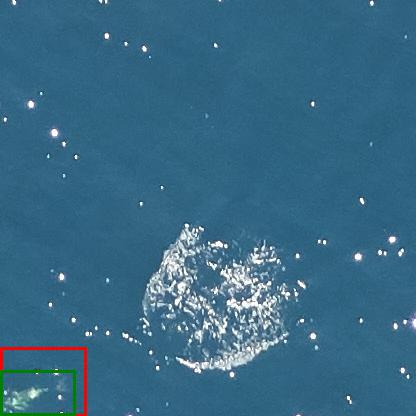}}; 
\end{tikzpicture}
}
\caption{Some favorable cases processed by the VQ-VAE model from the Semmacape dataset.}
\label{fig:favorable_cases}
\end{figure*}
Table \ref{table:res} shows better behavior of our VQ-VAE model on the two marine animal image datasets compared to three reference methods \cite{berg2022weakly, defard2021padim, kim2021semi}. From the table, a high gain of F1-score is achieved for both datasets (about $12\%$ better than the second best method). It should be noted that since VQ-VAEs could provide highly accurate reconstructions, they reduce the noise in the residual images. Therefore, the number of false detections is reduced, which yields high precision but may affect the recall. We observe this behavior from the first table on the Semmacape data. 
\begin{table}[ht]
\centering
\begin{tabular}{l|ccc}
    \hline
        \textbf{Method} &\textbf{F1-score }
        & \textbf{Recall} & \textbf{Precision }\\ 
        \hline
        PaDiM~\cite{defard2021padim} & 0.383 & 0.434 & 0.343 \\
        OrthoAD~\cite{kim2021semi} & 0.458 & 0.373 & 0.594 \\
        PaDiM+NF~\cite{berg2022weakly} & 0.530 & {\bf 0.757} & 0.408  \\
        \hline
        Ours & {\bf 0.636} &  0.497 & {\bf 0.884}  \\
        \hline
    \end{tabular} \\
    (a) Results on the Semmacape dataset.\\
\vspace{5mm}
\begin{tabular}{l|ccc}
    \hline
        \textbf{Method} &\textbf{F1-score }
        & \textbf{Recall} & \textbf{Precision }  \\ 
        \hline
        PaDiM~\cite{defard2021padim} & 0.504 & 0.443 & 0.586  \\
        OrthoAD~\cite{kim2021semi} & 0.571 & 0.514 & 0.643 \\
        PaDiM+NF~\cite{berg2022weakly} & 0.568 & 0.559 & 0.578 \\
        \hline
        Ours & {\bf 0.726} &  {\bf 0.754} & {\bf 0.701} \\
        \hline
    \end{tabular} \\
    (b) Results on the Kelonia dataset.
\caption{Comparative results on two marine animal datasets.}
\label{table:res}
\end{table}

\begin{figure*}
\centering
\makebox[0.9\textwidth][c]{
\begin{tikzpicture}
    \node[label={[above,xshift=0em, yshift=-0.08cm]Input}] (a){\includegraphics[width=2.5cm]{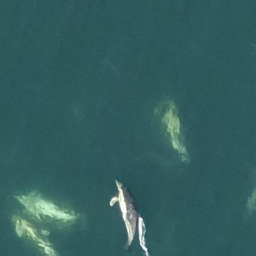}}; 
    \node[label={[above,xshift=0em, yshift=0cm]Reconstruction}](b) at (a.east) [xshift=11.2mm]{\includegraphics[width=2.5cm]{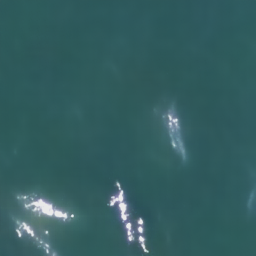}}; 
    \node[label={[above,xshift=0em, yshift=0cm]AM}](c) at (b.east) [xshift=11.2mm]{\includegraphics[width=2.5cm]{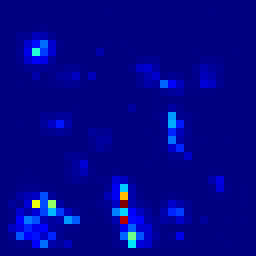}}; 
    \node[label={[above,xshift=0em, yshift=0cm]SM}](d) at (c.east) [xshift=11.2mm]{\includegraphics[width=2.5cm]{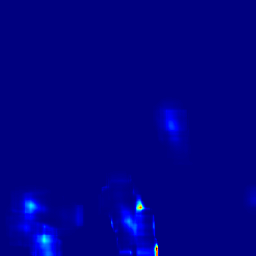}}; 
    \node[label={[above,xshift=0em, yshift=-0.08cm]Amap}](e) at (d.east) [xshift=11.2mm]{\includegraphics[width=2.5cm]{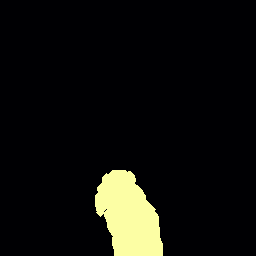}}; 
    \node[label={[above,xshift=0em, yshift=0cm]{\color{green}GT}  and {\color{red}Pred.}}](f) at (e.east) [xshift=11.2mm]{\includegraphics[width=2.5cm]{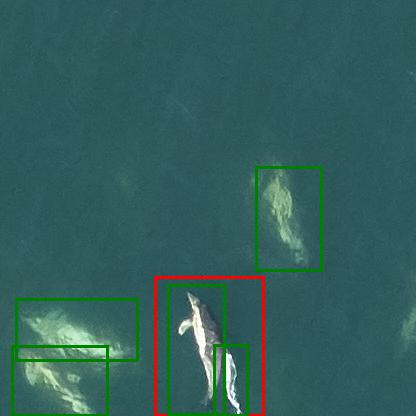}}; 
    \node at (a.south)[yshift=-11.5mm] {\includegraphics[width=2.5cm]{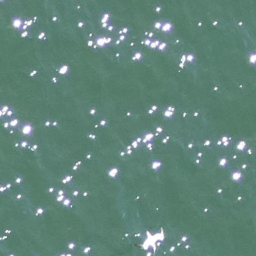}}; 
    \node at (b.south)[yshift=-11.5mm] {\includegraphics[width=2.5cm]{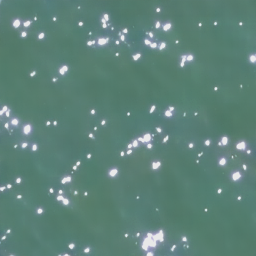}}; 
    \node at (c.south)[yshift=-11.5mm]{\includegraphics[width=2.5cm]{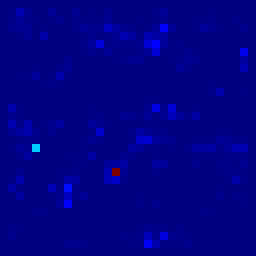}}; 
    \node at (d.south)[yshift=-11.5mm]{\includegraphics[width=2.5cm]{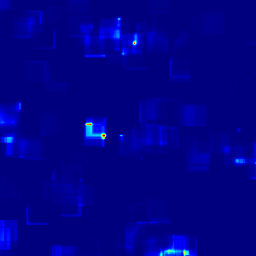}}; 
    \node at (e.south)[yshift=-11.5mm]{\includegraphics[width=2.5cm]{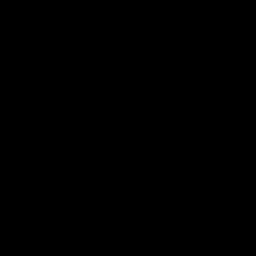}}; 
    \node at (f.south)[yshift=-11.5mm]{\includegraphics[width=2.5cm]{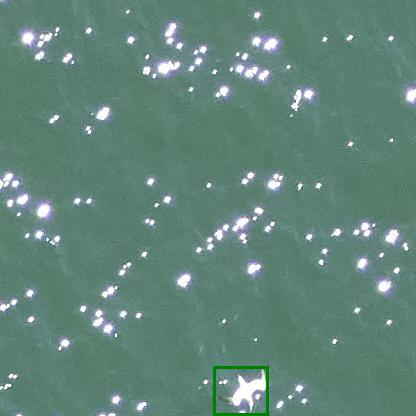}}; 
\end{tikzpicture}
}
\caption{Some complex cases processed by the VQ-VAE model from the Semmacape dataset.}
\label{fig:complex_cases}
\end{figure*}

We now provide some qualitative assessment on the use of VQ-VAE on the noisy Semmacape images. Fig. \ref{fig:favorable_cases} illustrates some favorable cases of good results yielded by the proposed method, while Fig. \ref{fig:complex_cases} shows some failed cases when dealing with different complex scenarios. 
First, it is interesting to notice from the figures that the model seems to reconstruct the anomalies (i.e., the animals) into sun glare or foam. This is an expected behaviour since the VQ-VAE has been trained without being presented any animal: it is thus unable to correctly reconstruct the latter. From Fig.~\ref{fig:favorable_cases}, apart from the simplest cases where the animal is isolated on an almost uniform background (as in the first row), the proposed approach also works well on usual cases where foam and sun glare appear and disturb the images. 
In Fig.~\ref{fig:complex_cases}, the first image is devoid of sun glare or foam but it depicts the case where dolphins are present at different distances from the sea surface. Our approach only detects 2 out of 5 dolphins. Then,
the second row illustrates the most complex scenario where images are corrupted by a lot of sun glare. In this case, the $SM$ map exhibits many false positives while the $AM$ is unable to correctly localize the anomalies. Thus, we end up with no detection at all after the double thresholding step.
\section{Conclusions}
\label{sec:4-Conclusions}
The context of AD enables us to take advantage of the huge amount of empty data which is devoid of interest in many traditional approaches. We have reported promising results in weakly-supervised detection of marine animals from aerial images by adapting the VQ-VAE model to the particularly noisy data. The VQ-VAE model learns a representation of the normality (empty images with no animal) during the training step. Then at prediction time, anomalies are spotted thanks to dedicated metrics and a double thresholding approach. 

One of the next steps in this unsupervised context using VQ-VAE architectures is to perform the counting and classification of the anomalies directly from from their discrete latent space with codebooks, without their explicit detection. This could lead to methodological and practical novelties that could be of crucial importance for the community. Future works may also focus on the use of self-supervised approach \cite{berg2022self, bauer2022self} to foster the representative capactity of VQ-VAE latent space.

\bibliographystyle{ieeetr}
\bibliography{refs}

\end{document}